\pdfoutput=1

\documentclass[11pt]{article}

\usepackage{EMNLP2022}

\usepackage{times}
\usepackage{latexsym}
\usepackage{multirow}
\usepackage{booktabs}
\usepackage{enumitem}
\usepackage{lipsum}
\usepackage{graphicx}
\usepackage{vcell}

\usepackage[T1]{fontenc}

\usepackage[utf8]{inputenc}

\usepackage{microtype}
\usepackage{makecell}
\usepackage{amsmath}
\DeclareMathOperator*{\argmax}{argmax} 

\makeatletter
\def\hlinewd#1{%
\noalign{\ifnum0=`}\fi\hrule \@height #1 %
\futurelet\reserved@a\@xhline}
\makeatother

\usepackage{cleveref}
\crefformat{section}{\S#2#1#3} 
\crefformat{subsection}{\S#2#1#3}
\crefformat{subsubsection}{\S#2#1#3}

\usepackage{tabularx}
\makeatletter
\def\hlinewd#1{%
\noalign{\ifnum0=`}\fi\hrule \@height #1 %
\futurelet\reserved@a\@xhline}
\makeatother

\title{Plug-and-Play Recipe Generation with Content Planning}

\author{
 \textbf{Yinhong Liu}$^\spadesuit$  \quad \quad
 \textbf{Yixuan Su}$^\spadesuit$  \quad \quad 
 \textbf{Ehsan Shareghi}$^{\heartsuit \spadesuit}$ \quad \quad 
 \textbf{Nigel Collier}$^\spadesuit$  \quad
 \\
 $^\spadesuit$Language Technology Lab, University of Cambridge\\
 $^\heartsuit$Department of Data Science and AI, Monash University\\
 {\tt \{yl535,ys484,nhc30\}@cam.ac.uk}\\
 {\tt ehsan.shareghi@monash.edu}
}

\begin{document}
\maketitle
\begin{abstract}
Recent pre-trained language models have shown promising capabilities in generating fluent and realistic natural language text.
However, generating multi-sentence text with global content planning has been a long-existing research question. 
Current approaches for controlled text generation can hardly address this issue, as they usually condition on \textit{single} known control attributes. 
In this study, we propose a low-cost yet effective framework which explicitly models the global content plan of the generated text. Specifically, it optimizes the joint distribution of the natural language sequence and the global content plan in a \textit{plug-and-play} manner. We conduct extensive experiments on the well-established Recipe1M+ benchmark. Both automatic and human evaluations verify that our model 
achieves the state-of-the-art performance on the task of recipe generation.\footnote{Our code and other related resources are publicly available at \url{https://github.com/williamLyh/RecipeWithPlans}.}

\end{abstract}

\section{Introduction}
Recent progress in large-scale language model pre-training has facilitated  significant improvement in generating increasingly realistic natural language text. Although this has been achieved on the surface-level fluency, it has been pointed that generating multi-sentence text with global constraints, or long-term planning is still far from being solved. Typical examples of such task include story continuation with logical coherency~\citep{nye2021improving,sinha2019clutrr}, and recipe generation with step-by-step planning~\citep{marin2019recipe1m+}. 

As suggested by~\citet{lecun2022path}, the aforementioned issues cannot be ameliorated by simply increasing the size of model parameters or the scale of pre-training data. 
Adding to this, current approaches for controlled text generation cannot directly tackle those issues either. 
For example, CTRL~\citep{keskar2019ctrl}, which trains a class-conditional language model, and PPLM~\cite{dathathri2019plug}, which re-ranks the language model predictions by an attribute model.
They usually share a common setup of optimizing conditional distributions $P(\boldsymbol{y}|a)$, where $\boldsymbol{y}$ is the text sequence and $a$ is the desired single control attribute. 
Examples of control attribute include sentiment \citep{ghosh2017affect}, topic \citep{tang2019topic} and formality \citep{wang2019topic}. 
However, content planning requires controlling with consideration of global context, which is more sophisticated than the single control attributes. 
Therefore, we identify the research gap for the current controlled text generation models to generate multi-sentence text with long-term content planning.


Motivated by previous research in cognitive science~\citep{evans2003two}, \citet{nye2021improving} pointed out that the reasoning of a neural-based model should consist of two systems, i.e. the \textit{system 1} makes intuitive and associative responses, and the \textit{system 2} makes deliberative and logical decisions. With greatly increased capabilities, large language models have become sufficiently competent to act as the system 1. However, we argue that, to address the aforementioned research gap, it is vital to empower the language models with the ability to make logical decisions, i.e. predict content plans. 

In contrast to the existing methods 
that optimize the conditional distributions,
we propose a novel framework which explicitly models the content plan $\boldsymbol{c}$ and optimizes the joint distribution $P(\boldsymbol{y}, \boldsymbol{c})$ in a \textit{plug-and-play} manner. Figure \ref{graph:overview} depicts an overview of our approach. Specifically, our proposed framework consists of (i) a content planner which predicts the global content plan of the output text; and (ii) a sequence generator, based on pre-trained language models, that generates the output following the content plan. 
The predicted content plan steers the generation of the sequence generator through a lightweight and plug-and-play style plan classifier.
It worth emphasizing that the sequence generator does not need to be trained with plan-specific data, which means adapting our framework to other Natural Language Generation (NLG) tasks is cheap and efficient.

We comprehensively evaluate our approach on the recipe generation task using the widely-used Recipe1M+ benchmark~\citep{marin2019recipe1m+}.
The experimental results demonstrate that our approach significantly outperforms previous state-of-the-art (SOTA) as judged by both automatic and human evaluations. 
In particular, the results show that the recipes generated by our model are more accurate and highly controllable. 

In summary, we conclude our contributions as two-fold: Firstly, we identify the current research gap and propose a novel framework that generates text with content plans in a plug-and-play manner.
Secondly, we conduct extensive experiments to show that our framework achieves SOTA performance on the widely-used Recipe1M+ benchmark.

\section{Background and Related Works}

\subsection{Controlled Text Generation}
Controlled Text Generation (CTG) refers to tasks of generating natural text conditioned on given controlled attributes. CTG approaches that leverage transformer-based Pre-trained Language Model (PLM) could be classified into three categories according to their required computation resources~\cite{zhang2022survey}. We provide a brief overview of these three categories.

\paragraph{Retraining.} These methods usually modify the original architecture of PLMs and retrain them for a specific downstream task. 
For example, CTRL, proposed by \citet{keskar2019ctrl}, is a representative that  trains a language model with task-specific control codes for each type of text corpus. 
Another work is POINTER, proposed by \citet{zhang2020pointer}, which stacks the architecture of insertion-based transformer \citep{chan2019kermit} in a hierarchical manner to enforce hard lexical constrains during text generation.
This type of methods could control the generated text effectively, but may negatively affect generalization of the PLM. They also usually have high computational footprint, and large-scale task-specific annotated data.

\paragraph{Fine-tuning.} These methods require partial or full fine-tuning of a PLM for each individual target attribute. 
For example, \citet{bostrom2021flexible} proposed ParaPattern which fine-tunes BART-based models \citep{lewis-etal-2020-bart} to generate text via applying different logical operations to premise inputs. 
\citet{ribeiro2021investigating} fine-tunes PLMs to control the generation from different types of graphical data.
The prefix-tuning, proposed by \citet{li2021prefix}, only optimizes a task-specific vector (prefix), while freezing the rest of PLM, to control the domain of generation. 
Fine-tuning PLMs based on a small amount of labelled data for the specific downstream task has achieved competitive performance.
However, fine-tuning based methods usually steer the PLM from the side of input, which means it could be hard to enforce hard constrains on the outputs directly.

\paragraph{Post-processing.} These methods usually do not require task-specific data to fine-tune the PLM, but require decoding algorithms to re-rank the generated text in a post-processing manner. 
As a representative work, PPLM, proposed by \citet{dathathri2019plug}, uses gradients from an attribute discriminant model to steer the text generation. 
FUDGE, proposed by \citet{yang2021fudge}, weights the decoding probabilities with an attribute predictor which takes partial sequence as input. 
\citet{su2022contrastive,su2022contrastiveiswhatyouneed} proposed Contrastive Search decoding, which encourages diversity by penalizing repetitive tokens.
\citet{lu2021neurologic} proposed NeuroLogic Decoding, which enforces the generation to satisfy a set of pre-defined hard lexical constrains.
MAGIC, proposed by \citet{su2022language}, applies an image relevance discriminator to guide the generation process with visual information. 
This type of methods are usually computationally cheap and flexible, because they have a separate guiding module. The increasing number of parameters of the PLM would not affect the complexities of the methods.
Our approach falls into this category of methods.

\subsection{Generation with Plan}
From the perspective of CTG tasks, attributes to control during generation include sentiment, topic, style, formality, story structure, content plan, among others. 
For example, \citet{ghosh2017affect} proposed Affect-LM, which extend an LSTM language model by conditioning on pre-defined affect categories and strength.
\citet{fu2018style} investigated the task of learning paper-news title style transfer from non-parallel data. 
\citet{li2020rigid} proposed the framework of SongNet which studies rigid format control to generate poems or songs that obey pre-defined rhyming schemes.

Previous works in controlling the generation with content planning are mainly focusing on the task of data-to-text generation and always taking \emph{schema selection} and \emph{ordering} as content plans.\footnote{Schema selection and ordering depend on input data structure, e.g. selecting and re-ordering the cells of tabular data or the nodes of graphical data.}
For example, \citet{moryossef-etal-2019-step} separates planning from neural text realization and takes the most probable traversal of RDF graph trees as content plan.
\citet{zhao-etal-2020-bridging} employs a GCN encoder to order the nodes of input RDF data as content plan.
\citet{su2021plan} proposed Plan-then-Generate, which treats orders of tabular schema as plans and then plans are encoded along with linearized data as inputs to a generative model. 
However, those methods require graphical or tabular input data and can only model content planning based on the data schema, which limits their domain of application.
\citet{yao2019plan} proposed a hierarchical generation framework that first plans a keyword storyline and then generate a story based on the storyline.
However, the generated keywords do not capture any global relation between each other.

\subsection{Recipe Generation}
Recipe generation refers to the task of generating recipe instructions from food images or textual ingredients and food title.
Because recipes have the natural step-by-step sequence flow, sentence-level content planning is desired in order to generate high quality recipes. 

Previous works tackled this issue in many directions. 
\citet{chandu-etal-2019-storyboarding} treated this problem as a Visual Story Telling task and built a dataset containing images and text for each intermediate step. The recipe instructions are generated from the sequential images. 
\citet{kiddon-etal-2016-globally} models global coherence of the recipes by maintaing an ingredient checklist dynamically. During generation, a language model is encouraged to refer to the checklist item.
\citet{bosselut2018simulating} tracks ingredient entity with a recurrent memory module and explicitly models actions as a set of per-defined state transforming operations. The recipes are then interpreted as structured collections of ingredient entities executed upon by cooking actions.
\cite{majumder2019generating} investigated the task of personalized recipe generation. The user's previously consumed recipes are encoded and attended by recipe name and ingredients to generate complete instructions.
However, they require complicated input data format, and sophisticated planning templates. 

Recipe1M+, introduced by \citet{marin2019recipe1m+}, is an extension of Recipe1M~\citep{Salvador_2017_CVPR} and contains over 1M textual recipes and ingredients and 13M corresponding food images. 
The dataset has been used for versatile tasks, such as image-recipe retrieval \citep{chen2017cross}, multi-modal embedding learning \citep{min2017delicious}, and recipe generation \citep{Salvador_2019_CVPR}. The RecipeGPT, proposed by \citet{h2020recipegpt}, finetuned a GPT-2 as a backbone generation model, taking only recipe titles and ingredients as input and recipe instructions as output. 
The NeuroLogic Decoding \citep{lu2021neurologic} takes the same setup, while enforcing hard lexical constrains on the occurrence of the ingredients. We follow the setup of these works and only consider the textual components of the Recipe1M+.

To the best of our knowledge, for the task of CTG with content planning, there has been no previous attempt neither on dataset with more flexible format such as recipe, nor with a plug-and-play post-process method.


\begin{figure*} [t]
\begin{center}
\includegraphics[width=\linewidth]{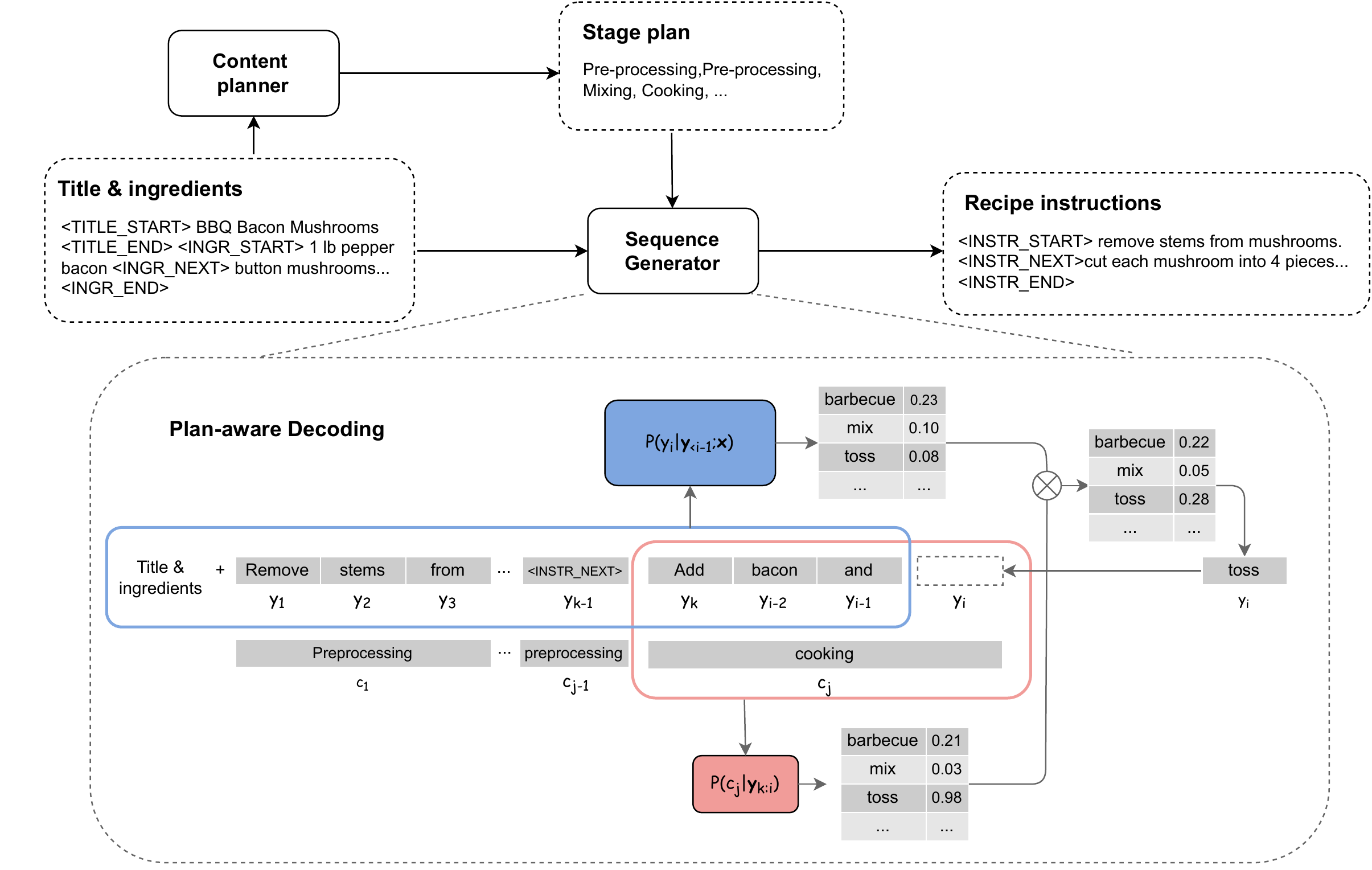}%
\end{center}
\caption{Model overview. The upper half demonstrates our framework.  Firstly, the title and ingredients are used to predict the stage plan by the content planner module. Then, the sequence generator module, guided by the stage plan, generates the recipe instructions.  
The bottom half illustrates one step of the plan-aware decoding. In the given example, the current stage is `cooking'. The language model (blue) outputs unconditional probabilities based on all previous context and inputs. The stage classifier (red) computes the probabilities of the current partial sentences belonging to the current stage `cooking'.}
\label{graph:overview}
\end{figure*}

\section{Methodology}
\subsection{Overview}
Figure \ref{graph:overview} depicts our proposed framework. Given the recipe title and ingredients, the content planner (\cref{sec:content plan}) first predicts the most probable content plan. The predicted content plan then guides the generation of the sequence generator (\cref{decoding}) via a lightweight and plug-and-play operation.  Below, we elaborate the details of the proposed approach.

\subsection{Content Planner}
\label{sec:content plan}
A carefully designed plan schema is vital for systems that require sophisticated controls. 
By examining recipe instructions, we observe the fact that they share a common structure of sequence of step-by-step stages and there are natural patterns behind those stage sequences. 
Therefore, we treat a content plan as a sequence of stages, where some stages could be of the same type.
Specifically, we define 7 types of instruction stage based on the processing step of the food, including:

\begin{itemize}[noitemsep,topsep=1pt]
    \itemsep 0em
    \item \textbf{Pre-processing} means the preparations of ingredients or cooker.
    \item \textbf{Mixing} includes actions of combining one or more ingredients together.
    \item \textbf{Transferring} is for the actions of moving or transferring food or intermediate food to a specific place.
    \item \textbf{Cooking} represents the actual cooking actions, which could vary drastically across different recipes.
    \item \textbf{Post-processing} usually refers to the following up actions after the `cooking' stage, such as `cooling down', `garnish'.
    \item \textbf{Final} refers to the last few actions before serving the food or the serving action itself.
    \item \textbf{General} includes the rest of actions which cannot be classified into the above categories.
\end{itemize}

\begin{table}[t]
    \small
	\centering  
	\renewcommand{\arraystretch}{1.2}
	\setlength{\tabcolsep}{6pt}
	\scalebox{1.1}{
	\begin{tabular}{ll}
	    \hlinewd{0.75pt}
	    \textbf{Stage Types}&\multicolumn{1}{l}{\textbf{Keywords}}\\
	    \hline
	    Pre-processing&Peel, beat, rinse, prepare ...\\
	    Mixing&Mix, add, combine, blend ...\\
	    Transferring&Move, put, pour, place ...\\
	    Cooking &Fry, bake, cook, boil, grill ...\\
	    Post-processing&Cool, shake, garnish, cover ...\\
	    Final &Serve, yield, wrap, enjoy ...\\
	    General&Uncovered or ambiguous verbs\\
		\hlinewd{0.75pt}
	\end{tabular}}
    \caption{Seven stage types and example keywords for each stage type.}
    	\vspace{-1.5mm}
	\label{tab:stage}
\end{table}

As recipe instructions are usually sentences led by action verbs, an assumption is made that the stage types of the instructions are decided by their main action verbs. For each type of stage, we assign a set of exclusive stage-specific action verbs, as shown in Table \ref{tab:stage}. For example, the `cooking' stage includes actions such as `fry', `bake', `boil', etc.
Then, we built a rule-based system that automatically tags recipe instructions with stage labels according to the pre-defined verb sets.
We tag the instructions from train set of Recipe1M+, which contains around 710K recipes, with the stage labels and refer to them as \textit{silver labels}.
\footnote{In Appendix \ref{sec:appendix:tagging}, we elaborate more implementation details of the rule-based stage tagging system.
In \cref{sec:classifier}, we evaluate the quality of the silver labels with human annotations on an evaluation subset.}
By this way, we can obtain the content plan of a recipe, i.e. the sequence of the stage labels.


After acquiring the content plan $\boldsymbol{c}=\{c_1, c_2, ..., c_{|\boldsymbol{c}|}\}$ using our rule-based system, the distribution of $\boldsymbol{c}$ is then modelled by the content planner as $P(\boldsymbol{c}|\boldsymbol{x})$, where $\boldsymbol{x}$ is the given recipe title as well as ingredients, and $c_j$ belongs to one of the seven stage types shown in Table \ref{tab:stage}. Specifically, given the recipe title and ingredients, we use a Seq2seq model, i.e. BART~\citep{lewis2020bart}, to model the content plan as 
\begin{equation}
    P(\boldsymbol{c}|\boldsymbol{x}) = \prod_{j=1}^{|\boldsymbol{c}|}p_{\theta_{c}}(c_j|\boldsymbol{c}_{<j}; \boldsymbol{x}),
\end{equation}
where $\theta_c$ is the parameters of the content planner. The main assumption of our modelling choice is that the content plan, i.e. cooking procedure, could be mostly determined once the target food and ingredients are known.




\subsection{Plan-Aware Decoding}
\label{decoding}
Given the recipe title and ingredients $\boldsymbol{x}$, and the content plan $\boldsymbol{c}$, we formulate the conditional distribution of recipe $\boldsymbol{y}$ by following the Bayes rule as

\begin{equation}
\begin{split}
	& P(\boldsymbol{y}|\boldsymbol{x}, \boldsymbol{c}) = \prod_{i=1}^{|\boldsymbol{y}|}p(y_i|\boldsymbol{y}_{<i}; \boldsymbol{x}, \boldsymbol{c})  \\
    & \propto \prod_{i=1}^{|\boldsymbol{y}|} p_{\theta_g}(y_i |\boldsymbol{y}_{<i}; \boldsymbol{x})\cdot p_{\theta_f}(c_j|\boldsymbol{y}_{k:i}),
\end{split}
\label{eq:generation}
\end{equation}
where $c_j$ refers to the stage that the current partial sequence $\boldsymbol{y}_{k:i}$ belongs to. The $\theta_f$ is an \textit{off-the-shelf} stage classifier which predicts the probability distribution over 7 stage classes by taking the partial sequence $\boldsymbol{y}_{k:i}$ as input. 
It should be noted that we assume the probability of the current stage label should \textit{only} depend on the partial sequence that belongs to the current stage.

During inference, based on Equation \ref{eq:generation}, the selection of the output token $\hat{y}_i$ at step $i$ follows

\begin{equation}
    \hat{y}_i =  \argmax_{y_i\in V_S} p_{\theta_g}(y_i |\boldsymbol{y}_{<i}; \boldsymbol{x})^{(1-\alpha)}\cdot p_{\theta_f}(c_j|\boldsymbol{y}_{k:i})^{\alpha},
\end{equation}
where $\alpha$ is a hyper-parameter that regulates the importance of two terms. 
$V_S$ is the set of top-$S$ predictions from the sequence generator's probability $p_{\theta_g}(\cdot |\boldsymbol{y}_{<i}; \boldsymbol{x})$ and $S$ is set as $5$ by default. 
We use the sequence generator's predictions on subset $V_S$ to approximate the predictions over the total vocabulary.
With this approximation, the stage classifier only needs to be applied upon $S$ candidates, therefore assuring the computational efficiency.

In this work, we fine-tune a GPT-2 model~\citep{radford2019language} on the training set of the Recipe1M+ benchmark to make it the sequence generator. To acquire the stage classifier, we fine-tune a lightweight DistilBERT~\citep{sanh2019distilbert} on the partial recipe instructions with the silver stage labels that we obtain as described in \cref{sec:content plan}.

Intuitively, our approach can be deemed as utilizing the stage classifier as a re-ranking step on the top $S$ candidates predicted by the sequence generator.
Figure \ref{graph:overview} illustrates an example, in which the sequence generator first predicts probabilities across all the vocabulary and the word `barbeque' has the highest likelihood. Then, the stage classifier re-ranks the predictions based on the current stage label `cooking' and assigns the highest probability to `toss'.

We note that using a partial sequence stage classifier to guide the decoding shares a similar idea with the previous study, i.e. FUDGE~\citep{yang2021fudge}. However, in contrast to FUDGE, our approach works on discriminating 7-class planning stages rather than only supporting binary attributes. In addition, to ensure the structural fluency of the generated recipe, we also control the generation from the perspective of global content planning, rather than focusing on one single control attribute.

\subsection{Advantages and Limitations}
\label{sec:limitations}
In this section, we discuss the theoretical advantages and limitations of our proposed approach.

We highlight the advantages that:
(i) We control the generation process in the plug-and-play manner, without the need of fine-tuning the language model, i.e. sequence generator module, with plan-specific data. In other words, given an off-the-shelf stage classifier and content planner, our framework is training-free.   
(ii) The stage classifier and content planner are both lightweight models compared to the sequence generator and can be fine-tuned with non-parallel data. 
(iii) Because the content plan schema is designed by humans, our framework can effectively inject human knowledge of the constrain patterns explicitly into the generation process.

We also point out the limitations of our framework:
(i) The overall performance of our model depends on the manually designed plan schema, which cannot be perfect, as it is based on heuristic experience. 
There are many cases where the stage of the instructions could be ambiguous.
For example, in a real recipe, it is possible for `slice the steak' to be both type of `pre-processing' or `post-processing'.
It is hard for humans to decide whether `add salt and pepper' is a special case of `seasoning', which belongs to the stage `pre-processing' or `mixing'.  
Another example is `pour milk and mix well', which contains two verbs from two stages.
(ii) As pointed out by \citet{zhang2022survey}, guided re-ranking algorithm, as a method of controlled text generation based on post-processing, suffers the problem of relatively low control strength, compared with the methods based on fine-tuning or retraining.

\section{Experiments}

In this section, we evaluate our method from three aspects: (i) The performance of planner module; (ii) the performance of the stage classifier; and (iii) the performance of the recipe generation.
The implementation details on these three parts are explained in \cref{sec:planner evaluation}, \cref{sec:classifier}, and \cref{sec: CTG evaluation}, respectively. In Appendix \ref{sec:appendix:case}, we provide examples that compare the generated results from our model and the baselines.

We pre-processed the Recipe1M+ dataset by firstly filtering out instructions that contain less than 3 words, e.g. 'combine all', as they are usually trivial. We also truncate recipes with too many instructions at the length of 15, because recipes with too many instructions usually include irrelevant information due to data scraping errors. 
By this way, about $9\%$ of the original Recipe1M+ are filtered out and the resulting dataset are used in our experiments.

\subsection{Planner Evaluation}
\label{sec:planner evaluation}

As described in~\cref{sec:content plan}, the content planner predicts the sequence of stage plans from the given recipe title and ingredients.
We took the sequences of the silver stage labels as reference plans and finetuned a seq2seq model, i.e. BART base version~\citep{lewis2020bart}. The silver labels are generated through the automatic tagging system described in \cref{sec:content plan}.
We evaluate the content planner module on the test set of the Recipe 1M+.

\begin{table}
\centering
\begin{tabular}{lc} 
\hlinewd{0.75pt}
\textbf{Metrics} & \textbf{Planner}        \\ 
\hline
Uni-gram         &         69.4                \\
Bi-gram          &            42.3             \\
Tri-gram         &               16.9          \\
Exact match      &             39.0  \\
\hlinewd{0.75pt}

\end{tabular}
\caption{Planner module evaluation results. Match rate accuracy (in percentage) for uni-gram, bi-gram, tri-gram and exact match, between predicted and reference plans}
\label{tab:planner}
\end{table}

\begin{figure} []
\begin{center}
\includegraphics[width=1\linewidth]{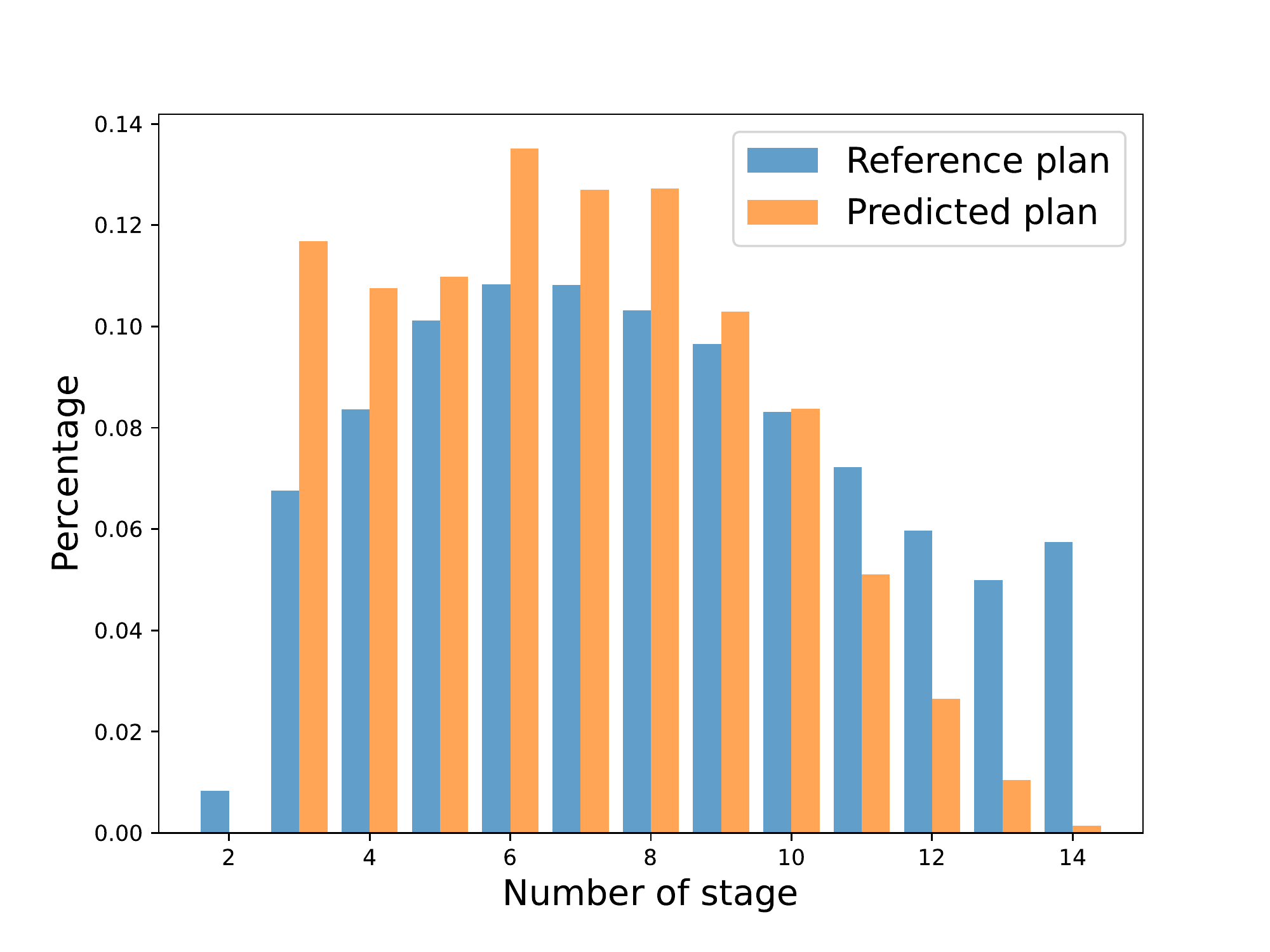}%
\end{center}
\caption{Histograms of the length of the predicted and the corresponding reference plans.}
\label{graph:lengths}
\end{figure}

Table~\ref{tab:planner} presents the evaluation results on the content planner, where the exact match rate is the percentage of matched stages in their exact positions of the reference plan. From the results, we see that our planner module achieves $39\%$ accuracy. 
Additional to this, because we are comparing two plan sequences, n-gram match rates are also important indicators to measure how good underlying patterns are learnt.
We show that for uni-gram and bi-gram we achieved relatively high match rates at $69.4\%$ and $42.3\%$, respectively; For tri-gram, we got  $16.9\%$ accuracy. 
This performance drop shows that our planner can learn the patterns between two successive instructions to an acceptable level, but the patterns among three successive instructions become hard to predict.

To further illustrate the performance of the content planner module, we also compare the distribution of lengths of the predicted and reference stage plans. 
As shown in Figure \ref{graph:lengths}, their histograms show similar bell-shape and the percentage of their mismatching is around $29.1\%$, which we consider as acceptably low. 
The main source of this, we believe, is due to the heavy tail of the distribution at length of 15.
As explained in Appendix \ref{sec:special tokens}, this is caused by the truncation of recipe instructions during the pre-processing steps.  

\begin{table}[]
\centering
\begin{tabular}{lc}
\hlinewd{0.75pt}

\textbf{Model}   & \textbf{Accuracy} \\ \hline
Stage classifier &    56.3     \\
Silver label     & 62.7          \\ 
\hlinewd{0.75pt}
\end{tabular}
\caption{Stage classifier evaluation results. The accuracy (in percentage) of the predictions of our stage classifier and automatically tagged silver stage labels, compared with manually labeled gold annotations. }
\label{tab:classifier accuracy}
\end{table}

\begin {table*}[t!]
\centering
\begin{tabular}{lccccc} 
\hlinewd{0.75pt}
\textbf{ Model}      & \textbf{BLEU}$\uparrow$ & \textbf{ROUGE-L}$\uparrow$  & \textbf{Plan Match}$\uparrow$ & \textbf{Coverage}$\uparrow$ & \textbf{Extra}$\downarrow$  \\ 
\hline
RecipeGPT, top-k     & 11.5          & 34.8          & 26.0          & 59.0              & 24.0            \\
RecipeGPT, beam      & 12.2          & 37.1          & 24.0          & 63.1              & 21.9            \\
NeuroLogic           & 11.8          & 38.2          & 21.8          & \textbf{67.1}     & 22.5            \\ 
\hline
Our model & \textbf{13.9} & \textbf{39.1} & \textbf{40.6} & 65.4              & \textbf{20.7}   \\ 
Our model, oracle    & 14.3          & 40.6          & 39.2          & 65.8              & 22.0            \\
\hlinewd{0.75pt}

\end{tabular}
\caption{Experimental results for our models and the baselines. \textit{Oracle} version of our model represents the plan-aware decoding is guided by the reference plan.  All models are evaluated on the same evaluation subset. $\uparrow$ means the higher the better, and $\downarrow$ means the lower the better.}
\label{tab:results}
\end{table*}

\subsection{Stage Classifier Evaluation}
\label{sec:classifier}
As described in~\cref{decoding}, the stage classifier predicts the stage label for the given full or partial recipe instruction. 
In our experiments, we implement the stage classifier by fine-tuning a DistilBERT with partial instructions along with the corresponding silver stage labels.
The partial instructions can be obtained by truncating the instructions from the training set of the Recipe1M+ at random positions.  
The size of the resulting partial instructions training set is around 4.9M.  
Below, we evaluate the performance of our stage classifier.

We construct a evaluation set by randomly sampling 300 examples from the recipes instructions in the test set of Recipe1M+. 
Then, we ask three human annotators proficient in English to annotate the instructions with stage labels following our provided guidelines. 
The human annotations are referred as \textit{gold labels}. 
In Table \ref{tab:classifier accuracy}, we evaluate the stage classifier and the rule-based tagging system with the gold labelled evaluation set.
The stage classifier achieves accuracy of $56.3\%$, while its upper bound, the silver labels, has accuracy of $62\%$.
We consider this is an acceptable performance, because this is a 7-class classification problem and there is subjective understanding of the imperfect plan schema, as explained in \cref{sec:limitations}.

\begin{table}
\centering
\begin{tabular}{lcc} 
\hlinewd{0.75pt}
\textbf{Method}        & \textbf{Fluency} & \textbf{Quality}  \\ 
\hline
RecipeGPT, beam        & 4.07             & 0.68              \\
Our model         & \textbf{4.34}    & \textbf{0.88}     \\ 
\hline
NeuroLogic             & 3.87             & 0.59              \\
Our model         & \textbf{4.27}    & \textbf{0.85}     \\ 
\hline
Our model, oracle & \textbf{4.31}    & \textbf{0.81}     \\
Our model         & 4.28             & 0.76              \\
\hlinewd{0.75pt}
\end{tabular}
\caption{Human evaluations on the generated recipes. A/B test style pairwise comparisons.}
\label{tab:human}
\end{table}

\subsection{Recipe Generation}
\label{sec: CTG evaluation}
In this section, we evaluate our plan-aware decoding method with both automatic  and human evaluations, and compare the performance with two strong baseline methods.
The sequence generator of our model is a base version of GPT-2, finetuned on the training set of the Recipe1M+ dataset.  
We pre-processed the recipe data with special separation tokens, as shown in the example in Figure \ref{graph:overview}, and more details are provided in Appendix \ref{sec:special tokens}.
Both our sequence generator as well as the compared baselines are fine-tuned on the same processed data.

\subsubsection{Baselines}

\textbf{RecipeGPT},  proposed by \citet{h2020recipegpt}, fine-tuned a base version of GPT-2 with the training set of the Recipe1M+ dataset.
During generation, it employs two types of decoding methods, top-k sampling and beam search. 
We re-implement the RecipeGPT as a representative of finetune-based methods and set the sampling candidate number and the beam size as 5.

\textbf{NeuroLogic decoding}, proposed by \citet{lu2021neurologic}, is a post-processing method which can be applied to different generative models.
It tries to search for optimal output sequences that satisfy a set of pre-defined lexical constrains. 
The constrains enforce certain words to appear or not appear in the generated sequences.
In this work, we choose the base version of GPT-2 as the underlying generative model and set the constrains such that all ingredients from the inputs should appear in the generated sequences.
The beam size is set as 5.

\subsubsection{Metrics}
\label{sec:metrics}
\textbf{Automatic Metrics}.
We use two widely-used metrics to assess the surface-level accuracy of the generated result, including BLEU~\citep{papineni2002bleu} and ROUGE-L~\citep{lin2002manual}. 
To measure the controlling ability of different models, we measure the plan match rate, which is the average percentage of the stage plan of the generated recipes that agree with the input stage plan.
The stage plans of the generated recipes are also labelled by the rule-based stage tagging system described in \cref{sec:content plan}.
In Table \ref{tab:results}, we refer the plan match rate as Plan Match.

We also explicitly measures the average percentage of coverage of the given ingredients and the percentage of hallucinated ingredients.
In Table \ref{tab:results}, we refer to them as coverage and extra, respectively. .
The details of how they are calculated are provided in Appendix \ref{sec:appendix:coverage}

\textbf{Human evaluations.}
To make a similar comparison, we follow the same human evaluation setup as previous studies such as FUDGE and PPLM~\citep{yang2021fudge,dathathri2019plug}.
Specifically, we run A/B test style human evaluations to compare our model with the two baselines on the aspects of fluency and quality in a manner of one-to-one pairwise comparison.
For each comparison, the two compared models both generate recipes based on 100 randomly selected recipe title and ingredients. 
The evaluators were asked to rate the fluency, in Likert scale from 1 to 5, and the quality of the generated recipes.
For the quality, the evaluators need to decide which recipe can reproduce the food described by the given title (recipe A, recipe B, neither or both).
In Table \ref{tab:human}, we report quality as the percentage of the recipes that are labelled as able to reproduce the food.
The details of the human evaluators are described in Appendix \ref{sec:appendix:human}.

\subsubsection{Results}

We create an evaluation subset by randomly sampling 4000 examples from the test set of the Recipe1M+. 
All experiments are conducted on this same subset. Table~\ref{tab:results} presents the results of different methods averaged over 5 runs with different random seeds.

Apart from the the aforementioned baselines, we also evaluate the oracle version of our model, which takes the reference stage plans as the guidance. 
The  experimental results show that our model outperforms all compared baselines on all metrics except for the ingredient coverage.
The differences are statistically significant for BLEU, ROUGE-L and Plan Match as judged by Sign Test with $p<0.01$. 
For the percentage of hallucination ingredients, the difference is weakly significant ($p<0.1$).
The performance gains of our model on BLEU and ROUGE-L suggests that it can produce recipes with better surface-level similarities by injecting the knowledge of content plans.  
On the metric of ingredient coverage, NeuroLogic decoding achieves the best results as it explicitly priorities the hard constrain of occurrence of the ingredients over surface-level fluency. 
It is worth noting that our models, including the oracle version, generally achieve significantly higher plan match rate than all the compared baselines.
This verifies that our model can effectively control the generation process of the recipe by following the given content plans.

For the human evaluation, our model is compared with the RecipeGPT and NeuroLogic baselines in pair and outperforms them on both fluency and recipe quality.
In addition, we observe that, with the help of the stage plan, our model can produce much less repeated, irrelevant or redundant instructions. 
Furthermore, by explicitly conditioning on stage plans, the recipes generated by our model are considered of better quality, which means they are easier for human readers to follow successfully.\footnote{In Appendix \ref{sec:appendix:case}, we provide detailed examples to compare the generated results from different methods.} 
Lastly, the oracle version of our model achieves further improved performances, suggesting that better stage plans can effectively provide human readers with better reading experience and more helpful guidance.





\section{Conclusion}
In this work, we first identify the research gap of the current controlled text generation models to generate text with sentence-level content planning. Then we propose a framework that optimizes the joint distribution of the natural language sequence and the content plans in a lightweight as well as plug-and-play manner. Extensive automatic and human evaluations demonstrate that our model achieves a new state of the art on the recipe generation task and outperforms previous studies by significant margins. Lastly, we show that our model can generate recipes that are more accurate and controllable by following the guidance of explicit content plans.


\bibliography{anthology,custom}

\begin{thebibliography}{43}
\expandafter\ifx\csname natexlab\endcsname\relax\def\natexlab#1{#1}\fi

\bibitem[{Bosselut et~al.(2018)Bosselut, Levy, Holtzman, Ennis, Fox, and
  Choi}]{bosselut2018simulating}
Antoine Bosselut, Omer Levy, Ari Holtzman, Corin Ennis, Dieter Fox, and Yejin
  Choi. 2018.
\newblock Simulating action dynamics with neural process networks.
\newblock In \emph{International Conference on Learning Representations}.

\bibitem[{Bostrom et~al.(2021)Bostrom, Zhao, Chaudhuri, and
  Durrett}]{bostrom2021flexible}
Kaj Bostrom, Xinyu Zhao, Swarat Chaudhuri, and Greg Durrett. 2021.
\newblock Flexible generation of natural language deductions.
\newblock In \emph{Proceedings of the 2021 Conference on Empirical Methods in
  Natural Language Processing}, pages 6266--6278.

\bibitem[{Chan et~al.(2019)Chan, Kitaev, Guu, Stern, and
  Uszkoreit}]{chan2019kermit}
William Chan, Nikita Kitaev, Kelvin Guu, Mitchell Stern, and Jakob Uszkoreit.
  2019.
\newblock Kermit: Generative insertion-based modeling for sequences.
\newblock \emph{arXiv preprint arXiv:1906.01604}.

\bibitem[{Chandu et~al.(2019)Chandu, Nyberg, and
  Black}]{chandu-etal-2019-storyboarding}
Khyathi Chandu, Eric Nyberg, and Alan~W Black. 2019.
\newblock \href {https://doi.org/10.18653/v1/P19-1606} {Storyboarding of
  recipes: Grounded contextual generation}.
\newblock In \emph{Proceedings of the 57th Annual Meeting of the Association
  for Computational Linguistics}, pages 6040--6046, Florence, Italy.
  Association for Computational Linguistics.

\bibitem[{Chen et~al.(2017)Chen, Ngo, and Chua}]{chen2017cross}
Jing-jing Chen, Chong-Wah Ngo, and Tat-Seng Chua. 2017.
\newblock Cross-modal recipe retrieval with rich food attributes.
\newblock In \emph{Proceedings of the 25th ACM international conference on
  Multimedia}, pages 1771--1779.

\bibitem[{Dathathri et~al.(2019)Dathathri, Madotto, Lan, Hung, Frank, Molino,
  Yosinski, and Liu}]{dathathri2019plug}
Sumanth Dathathri, Andrea Madotto, Janice Lan, Jane Hung, Eric Frank, Piero
  Molino, Jason Yosinski, and Rosanne Liu. 2019.
\newblock Plug and play language models: A simple approach to controlled text
  generation.
\newblock In \emph{International Conference on Learning Representations}.

\bibitem[{Evans(2003)}]{evans2003two}
Jonathan St~BT Evans. 2003.
\newblock In two minds: dual-process accounts of reasoning.
\newblock \emph{Trends in cognitive sciences}, 7(10):454--459.

\bibitem[{Fu et~al.(2018)Fu, Tan, Peng, Zhao, and Yan}]{fu2018style}
Zhenxin Fu, Xiaoye Tan, Nanyun Peng, Dongyan Zhao, and Rui Yan. 2018.
\newblock Style transfer in text: Exploration and evaluation.
\newblock In \emph{Proceedings of the AAAI Conference on Artificial
  Intelligence}, volume~32.

\bibitem[{Ghosh et~al.(2017)Ghosh, Chollet, Laksana, Morency, and
  Scherer}]{ghosh2017affect}
Sayan Ghosh, Mathieu Chollet, Eugene Laksana, Louis-Philippe Morency, and
  Stefan Scherer. 2017.
\newblock Affect-lm: A neural language model for customizable affective text
  generation.
\newblock In \emph{Proceedings of the 55th Annual Meeting of the Association
  for Computational Linguistics (Volume 1: Long Papers)}, pages 634--642.

\bibitem[{H.~Lee et~al.(2020)H.~Lee, Shu, Achananuparp, Prasetyo, Liu, Lim, and
  Varshney}]{h2020recipegpt}
Helena H.~Lee, Ke~Shu, Palakorn Achananuparp, Philips~Kokoh Prasetyo, Yue Liu,
  Ee-Peng Lim, and Lav~R Varshney. 2020.
\newblock Recipegpt: Generative pre-training based cooking recipe generation
  and evaluation system.
\newblock In \emph{Companion Proceedings of the Web Conference 2020}, pages
  181--184.

\bibitem[{Honnibal and Montani(2017)}]{spacy2}
Matthew Honnibal and Ines Montani. 2017.
\newblock {spaCy 2}: Natural language understanding with {B}loom embeddings,
  convolutional neural networks and incremental parsing.
\newblock To appear.

\bibitem[{Keskar et~al.(2019)Keskar, McCann, Varshney, Xiong, and
  Socher}]{keskar2019ctrl}
Nitish~Shirish Keskar, Bryan McCann, Lav~R Varshney, Caiming Xiong, and Richard
  Socher. 2019.
\newblock Ctrl: A conditional transformer language model for controllable
  generation.
\newblock \emph{arXiv preprint arXiv:1909.05858}.

\bibitem[{Kiddon et~al.(2016)Kiddon, Zettlemoyer, and
  Choi}]{kiddon-etal-2016-globally}
Chlo{\'e} Kiddon, Luke Zettlemoyer, and Yejin Choi. 2016.
\newblock \href {https://doi.org/10.18653/v1/D16-1032} {Globally coherent text
  generation with neural checklist models}.
\newblock In \emph{Proceedings of the 2016 Conference on Empirical Methods in
  Natural Language Processing}, pages 329--339, Austin, Texas. Association for
  Computational Linguistics.

\bibitem[{LeCun(2022)}]{lecun2022path}
Yann LeCun. 2022.
\newblock A path towards autonomous machine intelligence version 0.9. 2,
  2022-06-27.

\bibitem[{Lewis et~al.(2020{\natexlab{a}})Lewis, Liu, Goyal, Ghazvininejad,
  Mohamed, Levy, Stoyanov, and Zettlemoyer}]{lewis-etal-2020-bart}
Mike Lewis, Yinhan Liu, Naman Goyal, Marjan Ghazvininejad, Abdelrahman Mohamed,
  Omer Levy, Veselin Stoyanov, and Luke Zettlemoyer. 2020{\natexlab{a}}.
\newblock \href {https://doi.org/10.18653/v1/2020.acl-main.703} {{BART}:
  Denoising sequence-to-sequence pre-training for natural language generation,
  translation, and comprehension}.
\newblock In \emph{Proceedings of the 58th Annual Meeting of the Association
  for Computational Linguistics}, pages 7871--7880, Online. Association for
  Computational Linguistics.

\bibitem[{Lewis et~al.(2020{\natexlab{b}})Lewis, Liu, Goyal, Ghazvininejad,
  Mohamed, Levy, Stoyanov, and Zettlemoyer}]{lewis2020bart}
Mike Lewis, Yinhan Liu, Naman Goyal, Marjan Ghazvininejad, Abdelrahman Mohamed,
  Omer Levy, Veselin Stoyanov, and Luke Zettlemoyer. 2020{\natexlab{b}}.
\newblock Bart: Denoising sequence-to-sequence pre-training for natural
  language generation, translation, and comprehension.
\newblock In \emph{Proceedings of the 58th Annual Meeting of the Association
  for Computational Linguistics}, pages 7871--7880.

\bibitem[{Li et~al.(2020)Li, Zhang, Liu, and Shi}]{li2020rigid}
Piji Li, Haisong Zhang, Xiaojiang Liu, and Shuming Shi. 2020.
\newblock Rigid formats controlled text generation.
\newblock In \emph{Proceedings of the 58th annual meeting of the association
  for computational linguistics}, pages 742--751.

\bibitem[{Li and Liang(2021)}]{li2021prefix}
Xiang~Lisa Li and Percy Liang. 2021.
\newblock Prefix-tuning: Optimizing continuous prompts for generation.
\newblock In \emph{Proceedings of the 59th Annual Meeting of the Association
  for Computational Linguistics and the 11th International Joint Conference on
  Natural Language Processing (Volume 1: Long Papers)}, pages 4582--4597.

\bibitem[{Lin and Hovy(2002)}]{lin2002manual}
Chin-Yew Lin and Eduard Hovy. 2002.
\newblock Manual and automatic evaluation of summaries.
\newblock In \emph{Proceedings of the ACL-02 Workshop on Automatic
  Summarization}, pages 45--51.

\bibitem[{Lu et~al.(2021)Lu, West, Zellers, Le~Bras, Bhagavatula, and
  Choi}]{lu2021neurologic}
Ximing Lu, Peter West, Rowan Zellers, Ronan Le~Bras, Chandra Bhagavatula, and
  Yejin Choi. 2021.
\newblock Neurologic decoding:(un) supervised neural text generation with
  predicate logic constraints.
\newblock In \emph{Proceedings of the 2021 Conference of the North American
  Chapter of the Association for Computational Linguistics: Human Language
  Technologies}, pages 4288--4299.

\bibitem[{Majumder et~al.(2019)Majumder, Li, Ni, and
  McAuley}]{majumder2019generating}
Bodhisattwa~Prasad Majumder, Shuyang Li, Jianmo Ni, and Julian McAuley. 2019.
\newblock Generating personalized recipes from historical user preferences.
\newblock In \emph{Proceedings of the 2019 Conference on Empirical Methods in
  Natural Language Processing and the 9th International Joint Conference on
  Natural Language Processing (EMNLP-IJCNLP)}, pages 5976--5982.

\bibitem[{Marin et~al.(2019)Marin, Biswas, Ofli, Hynes, Salvador, Aytar, Weber,
  and Torralba}]{marin2019recipe1m+}
Javier Marin, Aritro Biswas, Ferda Ofli, Nicholas Hynes, Amaia Salvador, Yusuf
  Aytar, Ingmar Weber, and Antonio Torralba. 2019.
\newblock Recipe1m+: A dataset for learning cross-modal embeddings for cooking
  recipes and food images.
\newblock \emph{IEEE transactions on pattern analysis and machine
  intelligence}, 43(1):187--203.

\bibitem[{Min et~al.(2017)Min, Jiang, Wang, Sang, and Mei}]{min2017delicious}
Weiqing Min, Shuqiang Jiang, Shuhui Wang, Jitao Sang, and Shuhuan Mei. 2017.
\newblock A delicious recipe analysis framework for exploring multi-modal
  recipes with various attributes.
\newblock In \emph{Proceedings of the 25th ACM international conference on
  Multimedia}, pages 402--410.

\bibitem[{Moryossef et~al.(2019)Moryossef, Goldberg, and
  Dagan}]{moryossef-etal-2019-step}
Amit Moryossef, Yoav Goldberg, and Ido Dagan. 2019.
\newblock \href {https://doi.org/10.18653/v1/N19-1236} {{S}tep-by-step:
  {S}eparating planning from realization in neural data-to-text generation}.
\newblock In \emph{Proceedings of the 2019 Conference of the North {A}merican
  Chapter of the Association for Computational Linguistics: Human Language
  Technologies, Volume 1 (Long and Short Papers)}, pages 2267--2277,
  Minneapolis, Minnesota. Association for Computational Linguistics.

\bibitem[{Nye et~al.(2021)Nye, Tessler, Tenenbaum, and Lake}]{nye2021improving}
Maxwell Nye, Michael Tessler, Josh Tenenbaum, and Brenden~M Lake. 2021.
\newblock Improving coherence and consistency in neural sequence models with
  dual-system, neuro-symbolic reasoning.
\newblock \emph{Advances in Neural Information Processing Systems},
  34:25192--25204.

\bibitem[{Papineni et~al.(2002)Papineni, Roukos, Ward, and
  Zhu}]{papineni2002bleu}
Kishore Papineni, Salim Roukos, Todd Ward, and Wei-Jing Zhu. 2002.
\newblock Bleu: a method for automatic evaluation of machine translation.
\newblock In \emph{Proceedings of the 40th annual meeting of the Association
  for Computational Linguistics}, pages 311--318.

\bibitem[{Radford et~al.()Radford, Wu, Child, Luan, Amodei, Sutskever
  et~al.}]{radford2019language}
Alec Radford, Jeffrey Wu, Rewon Child, David Luan, Dario Amodei, Ilya
  Sutskever, et~al.
\newblock Language models are unsupervised multitask learners.

\bibitem[{Ribeiro et~al.(2021)Ribeiro, Schmitt, Sch{\"u}tze, and
  Gurevych}]{ribeiro2021investigating}
Leonardo~FR Ribeiro, Martin Schmitt, Hinrich Sch{\"u}tze, and Iryna Gurevych.
  2021.
\newblock Investigating pretrained language models for graph-to-text
  generation.
\newblock In \emph{Proceedings of the 3rd Workshop on Natural Language
  Processing for Conversational AI}, pages 211--227.

\bibitem[{Salvador et~al.(2019)Salvador, Drozdzal, Giro-i Nieto, and
  Romero}]{Salvador_2019_CVPR}
Amaia Salvador, Michal Drozdzal, Xavier Giro-i Nieto, and Adriana Romero. 2019.
\newblock Inverse cooking: Recipe generation from food images.
\newblock In \emph{Proceedings of the IEEE/CVF Conference on Computer Vision
  and Pattern Recognition (CVPR)}.

\bibitem[{Salvador et~al.(2017)Salvador, Hynes, Aytar, Marin, Ofli, Weber, and
  Torralba}]{Salvador_2017_CVPR}
Amaia Salvador, Nicholas Hynes, Yusuf Aytar, Javier Marin, Ferda Ofli, Ingmar
  Weber, and Antonio Torralba. 2017.
\newblock Learning cross-modal embeddings for cooking recipes and food images.
\newblock In \emph{Proceedings of the IEEE Conference on Computer Vision and
  Pattern Recognition (CVPR)}.

\bibitem[{Sanh et~al.(2019)Sanh, Debut, Chaumond, and
  Wolf}]{sanh2019distilbert}
Victor Sanh, Lysandre Debut, Julien Chaumond, and Thomas Wolf. 2019.
\newblock Distilbert, a distilled version of bert: smaller, faster, cheaper and
  lighter.
\newblock \emph{arXiv preprint arXiv:1910.01108}.

\bibitem[{Sinha et~al.(2019)Sinha, Sodhani, Dong, Pineau, and
  Hamilton}]{sinha2019clutrr}
Koustuv Sinha, Shagun Sodhani, Jin Dong, Joelle Pineau, and William~L Hamilton.
  2019.
\newblock Clutrr: A diagnostic benchmark for inductive reasoning from text.
\newblock In \emph{Proceedings of the 2019 Conference on Empirical Methods in
  Natural Language Processing and the 9th International Joint Conference on
  Natural Language Processing (EMNLP-IJCNLP)}, pages 4506--4515.

\bibitem[{Su and Collier(2022)}]{su2022contrastiveiswhatyouneed}
Yixuan Su and Nigel Collier. 2022.
\newblock Contrastive search is what you need for neural text generation.
\newblock \emph{arXiv preprint arXiv:2210.14140}.

\bibitem[{Su et~al.(2022{\natexlab{a}})Su, Lan, Liu, Liu, Yogatama, Wang, Kong,
  and Collier}]{su2022language}
Yixuan Su, Tian Lan, Yahui Liu, Fangyu Liu, Dani Yogatama, Yan Wang, Lingpeng
  Kong, and Nigel Collier. 2022{\natexlab{a}}.
\newblock Language models can see: Plugging visual controls in text generation.
\newblock \emph{arXiv preprint arXiv:2205.02655}.

\bibitem[{Su et~al.(2022{\natexlab{b}})Su, Lan, Wang, Yogatama, Kong, and
  Collier}]{su2022contrastive}
Yixuan Su, Tian Lan, Yan Wang, Dani Yogatama, Lingpeng Kong, and Nigel Collier.
  2022{\natexlab{b}}.
\newblock A contrastive framework for neural text generation.
\newblock \emph{arXiv preprint arXiv:2202.06417}.

\bibitem[{Su et~al.(2021)Su, Vandyke, Wang, Fang, and Collier}]{su2021plan}
Yixuan Su, David Vandyke, Sihui Wang, Yimai Fang, and Nigel Collier. 2021.
\newblock Plan-then-generate: Controlled data-to-text generation via planning.
\newblock In \emph{Findings of the Association for Computational Linguistics:
  EMNLP 2021}, pages 895--909.

\bibitem[{Tang et~al.(2019)Tang, Li, and Jin}]{tang2019topic}
Hongyin Tang, Miao Li, and Beihong Jin. 2019.
\newblock A topic augmented text generation model: Joint learning of semantics
  and structural features.
\newblock In \emph{Proceedings of the 2019 Conference on Empirical Methods in
  Natural Language Processing and the 9th International Joint Conference on
  Natural Language Processing (EMNLP-IJCNLP)}, pages 5090--5099.

\bibitem[{Wang et~al.(2019)Wang, Gan, Xu, Zhang, Wang, Shen, Chen, and
  Carin}]{wang2019topic}
Wenlin Wang, Zhe Gan, Hongteng Xu, Ruiyi Zhang, Guoyin Wang, Dinghan Shen,
  Changyou Chen, and Lawrence Carin. 2019.
\newblock Topic-guided variational auto-encoder for text generation.
\newblock In \emph{Proceedings of the 2019 Conference of the North American
  Chapter of the Association for Computational Linguistics: Human Language
  Technologies, Volume 1 (Long and Short Papers)}, pages 166--177.

\bibitem[{Yang and Klein(2021)}]{yang2021fudge}
Kevin Yang and Dan Klein. 2021.
\newblock Fudge: Controlled text generation with future discriminators.
\newblock In \emph{Proceedings of the 2021 Conference of the North American
  Chapter of the Association for Computational Linguistics: Human Language
  Technologies}, pages 3511--3535.

\bibitem[{Yao et~al.(2019)Yao, Peng, Weischedel, Knight, Zhao, and
  Yan}]{yao2019plan}
Lili Yao, Nanyun Peng, Ralph Weischedel, Kevin Knight, Dongyan Zhao, and Rui
  Yan. 2019.
\newblock Plan-and-write: Towards better automatic storytelling.
\newblock In \emph{Proceedings of the AAAI Conference on Artificial
  Intelligence}, volume~33, pages 7378--7385.

\bibitem[{Zhang et~al.(2022)Zhang, Song, Li, Zhou, and Song}]{zhang2022survey}
Hanqing Zhang, Haolin Song, Shaoyu Li, Ming Zhou, and Dawei Song. 2022.
\newblock A survey of controllable text generation using transformer-based
  pre-trained language models.
\newblock \emph{arXiv preprint arXiv:2201.05337}.

\bibitem[{Zhang et~al.(2020)Zhang, Wang, Li, Gan, Brockett, and
  Dolan}]{zhang2020pointer}
Yizhe Zhang, Guoyin Wang, Chunyuan Li, Zhe Gan, Chris Brockett, and William~B
  Dolan. 2020.
\newblock Pointer: Constrained progressive text generation via insertion-based
  generative pre-training.
\newblock In \emph{Proceedings of the 2020 Conference on Empirical Methods in
  Natural Language Processing (EMNLP)}, pages 8649--8670.

\bibitem[{Zhao et~al.(2020)Zhao, Walker, and
  Chaturvedi}]{zhao-etal-2020-bridging}
Chao Zhao, Marilyn Walker, and Snigdha Chaturvedi. 2020.
\newblock \href {https://doi.org/10.18653/v1/2020.acl-main.224} {Bridging the
  structural gap between encoding and decoding for data-to-text generation}.
\newblock In \emph{Proceedings of the 58th Annual Meeting of the Association
  for Computational Linguistics}, pages 2481--2491, Online. Association for
  Computational Linguistics.

\end{thebibliography}
\bibliographystyle{acl_natbib}

\appendix

\section{Appendix}
\label{sec:appendix}
\subsection{Details of preprocessing}
\label{sec:special tokens}
We then further processed the recipes by adding special separation tokens, as shown in the example in Figure \ref{graph:overview}. 
The separation tokens include $<TITLE\_START>$ and $<TITLE\_END>$ to wrap the recipe title, $<INGR\_START>$, $<INGR\_END>$ and $<INGR\_NEXT>$ to wrap and split the recipe ingredients. 
Similarly, $<INSTR\_START>$, $<INSTR\_END>$ and $<INSTR\_NEXT>$ are used to wrap and split the recipe instructions. 
There is no leaking of the stage label information from the separation tokens. 

\subsection{Details of metrics computing}
\label{sec:appendix:coverage}
To identify the ingredients in the generated recipes, we first create a list of total input ingredients in the Recipe1M+ dataset and then identify the ingredients in the recipes by string match.
The hallucination percentage is the number of hallucinated ingredients over the total number of ingredients in the input.
Ingredients that are not included in the input, but included in the total ingredient list, are considered as hallucinated.
It is worth noting that because of the limitations of string match, which cannot deal with plural, quantifier, synonym and etc, the coverage and hallucination percentages are not perfect. 
Therefore, they are better interpreted as rough indicators and used to compare between models parallelly.

\subsection{Rule-based stage label tagging system}
\label{sec:appendix:tagging}
In this section, we elaborate how we implement the rule-based tagging system.
To process on instruction, firstly we use the tokenizer from Python package Spacy \citep{spacy2} to identify all the verbs by checking the Part-Of-Speech (POS) tag of each token. 
Then we remove the verbs that are in the clauses by identifying punctuation and conjunction words.
If, by this point, there are more than one verbs left, we always keep the first verb as the main verb and tag the instruction base on this verb.
We tag the stage label by looking up which stage type the main verb belongs to, as shown in the example keywords in Table \ref{tab:stage}.

\subsection{Details of Human evaluators}
\label{sec:appendix:human}
For the tasks of human annotator in Section \ref{sec:classifier} and evaluation in Section \ref{sec:metrics}, we ask three voluntary university students whose first language are English to conduct each task. 
They are paid with rate of 10 GBP per hour, which is above the minimum hourly wage in UK. 

\subsection{Case study: Recipe Generation Example}
\label{sec:appendix:case}
In Table \ref{tab:case_study}, we show generation examples of three recipes. Instructions labeled in red show undesired characteristics. 

In the first example, the `Eggs Barbados', the recipes generated by baselines, RecipeGPT and NeuroLogic, both have the instruction `Preheat oven to 350 degrees f.'. We consider this instruction is irrelevant because the oven is never used in the rest of the recipes.

In the second example, the `Our Favourite Wheat Bread', the recipe generated by RecipeGPT shows undesired repetitive patterns on the instructions `Punch down and let rise again.'.

In the third example, the `Pomegranate Ice Tea', the recipe generated by RecipeGPT is inaccurate by `Serve immediately', because it lacks the step of `Pour over ice' or `add ice'.

\begin{table*}[t]
    \small
	\centering  
	\renewcommand{\arraystretch}{1.2}
        \hskip-1cm
        \setlength{\tabcolsep}{6pt}
	\scalebox{0.88}{
	\begin{tabular}{lclc}
	    \hlinewd{0.75pt}
	    \textbf{Recipe Title}&\textbf{Ingredients}&\textbf{Model}&\textbf{Generated Results}\\
	    \hline
	    \multirow{19}{*}{\makecell[l]{Eggs Barbados}}&\multirow{19}{*}{\makecell[l]{
     1 bunch green onions;\\
     1 package cream cheese;\\
     1 tablespoon butter;\\
     12 whole eggs;\\
     1 tablespoon milk;\\
     salt and pepper to taste}}
     &RecipeGPT&\makecell[l]{
     {\color{red}Preheat oven to 350 degrees f.} Slice the green onions in half\\ lengthwise. Slice the cream cheese in half lengthwise. Melt\\ the butter in a large skillet over medium heat. Add the green\\ onions and saute until tender. Add the cream cheese, eggs,\\ milk and salt and pepper to taste. Cook until the eggs are set.\\ Serve immediately.}\\
	    \cline{3-4}
	    &&NeuroLogic&\makecell[l]{
     {\color{red}Preheat oven to 350 degrees f.} Slice the green onions and the\\ cream cheese in half lengthwise. Melt the butter in the pan\\ over medium heat. Add the green onions and saute until tender.\\ Add the cream cheese, eggs, milk and salt and pepper to taste.\\ Cook until the eggs are completely cooked.}\\
	    \cline{3-4}
	    &&Ours&\makecell[l]{
     Slice the green onions into 1/4 inch slices. Melt the butter in\\ a frying pan. Add the chopped onions and saute until they are\\ soft. Add the cream cheese and mix well. In a separate bowl,\\ beat the eggs and milk. Add the eggs to the cream cheese mixture\\ and mix well. Add the green onions and salt and pepper to taste.\\ Drop by spoonfuls onto a greased baking sheet. Bake at 350\\ degrees for 20 minutes.}\\
	    \cline{3-4}
	    &&Reference&\makecell[l]{
     Chop your green onions and break up the cream cheese into\\ little chunks. Melt the butter in your frying pan on medium-low\\ heat. While the butter melts, beat the eggs with the milk until\\ combined. Add your salt and pepper. pour the egg mixture\\ into the pan. Dont stir up the eggs too much, just occasionally\\ scrape the bottom to mix up the cooked egg with the runny stuff.\\ Once the eggs are mostly cooked but still look a little wet, turn\\ off the heat. You dont want overcooked eggs, theyre rubbery and\\ icky! As soon as you turn off the heat, mix the cream cheese\\ chunks and chopped green onions into the eggs. Let it sit for just\\ a few seconds, so that the heat of the pan melts the cream cheese.}\\
		\hlinewd{0.75pt}
  	    \multirow{14}{*}{\makecell[l]{Our Favourite\\ Wheat Bread}}&\multirow{14}{*}{\makecell[l]{1 tablespoon dry yeast;\\2 cups warm water; \\14 cup soft butter; \\18 cup molasses;\\14 cup honey;\\1 teaspoon salt; \\3 cups whole wheat flour;\\ 2 cups white flour.}}
       &RecipeGPT&\makecell[l]{
       Dissolve yeast in warm water. Add butter, molasses, honey, and\\ salt. Mix well. Add whole wheat flour and white flour. Knead\\ until smooth and elastic. Cover and let rise in a warm place for\\ 1 hour. {\color{red}Punch down and let rise again. Punch down again and} \\{\color{red}let rise again. Punch down again and let rise again. Punch down} \\{\color{red}again and let rise again. Punch down again and let rise again.} \\{\color{red}Punch down again and let rise again. Punch down again and} \\{\color{red}let rise again. Punch down again and let rise again}}\\
	    \cline{3-4}
	    &&NeuroLogic&\makecell[l]{
     Dissolve yeast in warm water. Add water, butter, molasses, honey, \\and salt. Mix well. Add whole wheat flour and white flour. Mix\\ well. Knead until smooth and elastic. Cover and let rise in a warm\\ place for 1 hour. Bake for 1 hour.}\\
	    \cline{3-4}
	    &&Ours&\makecell[l]{
     Dissolve yeast in warm water. Add butter, molasses, honey, and\\ salt. Add whole wheat flour and white flour. Mix until smooth.\\ Cover and allow to rise in a warm place until doubled in size.\\ Punch down and let rise again. Pre-heat oven to 350 degrees.\\ Grease a loaf pan. Pour in batter. Bake for 1 hour.}\\
	    \cline{3-4}
	    &&Reference&\makecell[l]{
     Dissolve yeast in warm water. Combine butter, molasses, honey,\\ and salt. Add yeast mixture. Gradually add flour. Turn onto\\ floured surface and knead until smooth. Place in greased bowl\\ and let rise until double. Let dough rest a few minutes. Shape into\\ 
     2 loaves. Place in greased bread pans. Let rise about 1 hour. Bake\\ at 375 for 35-40 minutes.}\\
		\hlinewd{0.75pt}
	    \multirow{7}{*}{\makecell[l]{Pomegranate Iced Tea}}&\multirow{7}{*}{\makecell[l]{
     2 ounces freshly squeezed\\
     pomegranate juice or\\
     2 ounces store-bought\\
     pomegranate juice; \\
    5 ounces tea; \\
    1 ounce simple syrup; \\
    lemon wedge.}} 
     &RecipeGPT&\makecell[l]{
     Combine the pomegranate juice, tea, simple syrup, and lemon \\wedge in a pitcher. {\color{red}Serve immediately.}}\\
	    \cline{3-4}
	    &&NeuroLogic&\makecell[l]{
     Combine the pomegranate juice, tea, simple syrup, and lemon \\wedge in a pitcher. Add ice and stir well.}\\
	    \cline{3-4}
	    &&Ours&\makecell[l]{
     Combine the pomegranate juice, tea, simple syrup and lemon \\wedge in a pitcher. Add ice and stir well. Pour into a chilled \\glass and garnish with a lemon wedge.}\\
	    \cline{3-4}
	    &&Reference&\makecell[l]{
     Prepare tea and cool. Mix the rest of the ingredients. Pour over \\ice in an iced tea glass. Garnish with lemon wedge.}\\
		\hlinewd{0.75pt}
  
	\end{tabular}}
    \caption{Case Study. Generation examples of 3 recipes by our model and baselines. {\color{red} Instructions labeled in red} are considered problematic. }
    	\vspace{-1.5mm}
	\label{tab:case_study}
\end{table*}


\end{document}